 \documentclass[pmlr,twocolumn,10pt]{jmlr} 

\usepackage{booktabs}
\usepackage{multirow}
\usepackage{listings}
\usepackage{algorithm}
\usepackage{algorithmic}
\usepackage{float}

\usepackage[load-configurations=version-1]{siunitx} 

\newcommand{\equal}[1]{{\hypersetup{linkcolor=black}\thanks{#1}}}

\newcommand{\doctor}{medical professional}

\DeclareMathOperator*{\argmax}{arg\,max}

\theorembodyfont{\upshape}
\theoremheaderfont{\scshape}
\theorempostheader{:}
\theoremsep{\newline}

\jmlrvolume{LEAVE UNSET}
\jmlryear{2021}
\jmlrsubmitted{LEAVE UNSET}
\jmlrpublished{LEAVE UNSET}
\jmlrworkshop{Machine Learning for Health (ML4H) 2021} 

\title{MEDCOD: A \textbf{M}edically-Accurate, \textbf{E}motive, \textbf{D}iverse, and \textbf{CO}ntrollable \textbf{D}ialog System}

\author{
  Rhys Compton\equal{NYU. Work done while research intern at Curai}  \Email{rc4499@nyu.edu}\\
  Ilya Valmianski \Email{ilya@curai.com}\\
  Li Deng \Email{li@curai.com}\\
  Costa Huang\equal{Drexel University. Work done while intern at Curai} \Email{costa.huang@outlook.com}\\
  Namit Katariya \Email{namit@curai.com}\\
  Xavier Amatriain \Email{xavier@curai.com} \\
  Anitha Kannan  \Email{anitha@curai.com} \\
  \addr Curai
}

\begin{document}

\maketitle
\begin{abstract}
\label{sec:abstract}
We present \textbf{MEDCOD}, a \textbf{M}edically-Accurate, \textbf{E}motive, \textbf{D}iverse,  and \textbf{Co}ntrollable \textbf{D}ialog system with a unique approach to the natural language generator module. MEDCOD has been developed and evaluated specifically for the history taking task. It integrates the advantage of a traditional modular approach to incorporate (medical) domain knowledge with modern deep learning techniques to generate flexible, human-like natural language expressions. Two key aspects of MEDCOD's natural language output are described in detail. First, the generated sentences are emotive and empathetic, similar to how a doctor would communicate to the patient.  Second, the generated sentence structures and phrasings are varied and diverse while maintaining medical consistency with the desired medical concept (provided by the dialogue manager module of MEDCOD). Experimental results demonstrate the effectiveness of our approach in creating a human-like medical dialogue system. Relevant code is available at \url{https://github.com/curai/curai-research/tree/main/MEDCOD}
\end{abstract}

\section{Introduction}
\label{sec:intro}

The development of natural language (NL) understanding and dialogue systems, both spoken and text-based, has over 30 years of history and can be divided into three generations according to the disparate styles of system design: (1) use of expert systems based on symbolic-rules and templates \citep{Allen96, rudnicky1999agenda}, (2) use of (shallow) statistical learning
\citep{wang2005spoken,SLUBook-chapter,SLUBook-chapter3}, 
and (3) use of deep learning \citep{Tur2018,dhingra2016end}. The earlier two generations of dialogue systems were usually designed with a number of separate modules: textual (or spoken) natural language understanding (NLU), dialogue manager, natural language generation (NLG), and (optionally) spoken language generation. The main advantage of modular designs is their ability to easily incorporate domain knowledge. The main disadvantage is their weakness in generating flexible, human-like responses. The third-generation language understanding and dialogue systems, driven by deep learning technology \citep{Hinton2012,Huang2013,Deng2014,Li2016,Deng2018BOOK}, adopt the end-to-end neural network architecture approach. This provides the opportunity to learn all parts of the dialogue system jointly and the ability to produce more human-like responses \citep{vinyals2015neural,chen2016end2end,Wu2020,Hosseini2020}. The main weakness, however, is that end-to-end learned systems require large amounts of training data to implicitly acquire domain knowledge and suffer from poor control over the system's output. 

In this paper, we present a hybrid modular and deep learning approach to designing a medical dialogue system targeted for the history taking task called \textbf{MEDCOD}, which integrates domain knowledge and controllability from a modular design with human-like NLG of a deep learning system. Medical dialogues between patients and doctors are one key source of information for diagnosis and decision making \citep{Chen2020, Soltau2021}. Notably, the task of history taking in such medical dialogues is an important, time consuming but, in many circumstances, low-complexity part of a medical encounter as it involves asking a series of closed-ended questions to ascertain the patient's current condition; this makes it a prime target for automation, decreasing the clerical load on physicians and allowing them to practice at the top of their scope. Other clinical use cases include AI-driven online symptom checkers and automatic patient triaging.

In our history-taking dialogue system, the dialogue manager uses both an expert system and a machine learned emotion classifier to control a deep-learning-based NLG module. The expert system uses a medical knowledge base (KB) ({\it c.f.}~\cite{miller1990quick}) that contains rich medical domain knowledge to identify which patient-reportable medical finding should be asked next. The emotion classifier then predicts the emotion with which the NLG module should ask the question. The NLG module is implemented using a deep learning approach to generate 
variable medical questions while maintaining medical consistency with the expert-system-derived finding, while containing emotion-classifier specified emotion.

The technical contributions of this paper are as follows. First, we developed a novel method of using ``control codes'' \citep{Keskar2019} to within the medical dialogue data for training DialoGPT (dialogue generative pre-trained transformer) \citep{zhang2019}, which serves as the NLG model in our dialogue system.  This use of control codes aims to maintain medical consistency in the generated questions while creating diversity that exhibits human-like attributes. Second, we train an emotion classifier for use in the inference stage of NLG. This gives our system the ability to generate emotive sentences simulating human doctors' behavior. Finally, to overcome the problem of sparsity in the dialogue training data, we made effective use of GPT-3 \citep{brown2020} to augment \verb|finding|-\verb|NL| paired data jointly for both diversity and emotion while maintaining medical consistency in the NL output. With these technical innovations, we have built \textbf{MEDCOD}, the first medically consistent and controllable history taking dialogue system with human-like NL expression as the system output in each dialogue turn.

\section{Related Work}
\label{sec:related_work}

\textbf{The task-oriented dialog system} is one of four major types of dialogue systems in common use, the other three types being for non-goal-oriented applications in information consumption, decision support, and social interactions; see a review in \citep{Celikyilmaz2018}. This paper is devoted to task-oriented systems only, for the task of medical history taking.


The classic task-oriented dialog systems incorporate several components including Speech Recognition, Language Understanding, Dialog Manager (consisting of State Tracker and Dialog Policy), NLG, and Speech Synthesis. In the preliminary development, \textbf{MEDCOD} has not yet incorporated the speech components, although this would have a positive impact on further usability \citep{HeLi2013,Huang2010,HeDengIcassp2011,Deng2003,Yu2015}.

Until recently, a majority of task-oriented dialogue systems were based primarily on hand-crafted rules \citep{Aust,Simpson} or (shallow) machine learning techniques for all major components of the systems \citep{Gorin}. This work formulated the dialogue as a sequential decision making problem based on Markov decision processes and reinforcement learning \citep{young2010hidden}. With the introduction of deep learning in speech recognition, spoken language understanding, and dialog modeling,
incredible successes were demonstrated in the robustness and coherency of dialog systems, especially in the NLU component of the system. \citep{tur2012towards,RNN-TASL,hakkani2016multi,Jiwei2016,Lipton}. 

The approach we take in developing our current medical dialogue system has been motivated by successes in the related work discussed above. Our work is also inspired by \citet{Paranjape2020}, who demonstrated an open-domain, non-target-oriented dialogue system capable of empathetic conversations with emotional tone. Further, we have benefited from the work of \citep{Keskar2019}, which proposed the use of ``control codes'' to construct a generic language model for controllable NLG. We have used a similar mechanism to train the NLG module of our domain-specific medical dialogue system to maintain medical accuracy while generating diverse NL sentences. Finally, the approach to few-shot NLG using structured knowledge by \citet{Chen2020b} relates to our work in that we have also made use of a specific medical knowledge base (KB) (as part of our dialogue manager) to provide control over the NLG model; one key difference is that our KB is used to create medical concepts for their diverse NL expressions rather than to provide better generalization across domains. 

\textbf{Medical dialogue systems} have been reported in the literature only in recent years, with focus on NLU for clinical documents. Prior to the work presented in this paper, however, there does not appear to be any previous work that focused on NLG for medical dialogue systems. \citet{Enarvi2020} developed a system to generate NL medical reports based on patient-doctor dialogue transcripts but they did so from speech recognition outputs instead of from the semantic representation of a dialogue manager as in our work presented here. Below we briefly review the related work on NLU, which can be considered as complementary work to our current focus. 

In the area of mapping extracted medical concepts in conversations to a knowledge base, \citep{du2019, Du2019b} introduced a hierarchical two-stage approach to infer clinical entities (e.g., symptoms), their properties (e.g., duration), and relationships between them.\citet{selvaraj2019} focused on the problem of treatment regimen extraction while  \citet{khosla20} studied the problem of extracting relevant task-relevant utterances from medical dialogues. The problem of intent detection in doctor-patient interactions is a new area and was recently explored in \citep{Rojowiec2020}. Further, \citet{Soltau2021} recently reported a spoken medical dialogue system aiming to extract clinically relevant information from medical conversations between doctors and patients. But the work did not proceed further to use the extracted information as input to a dialogue manager and then to produce the NL response, which our MEDCOD approach does. \citet{Bo2019} introduces a medical knowledge-based dialogue system that acts as a health assistant. \citet{Lin2020} explored a natural paradigm for low-resource medical dialogue generation using very small amount of data for adaptation. Finally, \citet{Liu2019} presented a domain-aware automatic chest X-ray radiology report generation system. All the above work benefited the design of the NLG module of our MEDCOD system.





\section{Approach}
\label{sec:model_approach}
\begin{figure}
    \centering
    \includegraphics[width=0.8\linewidth]{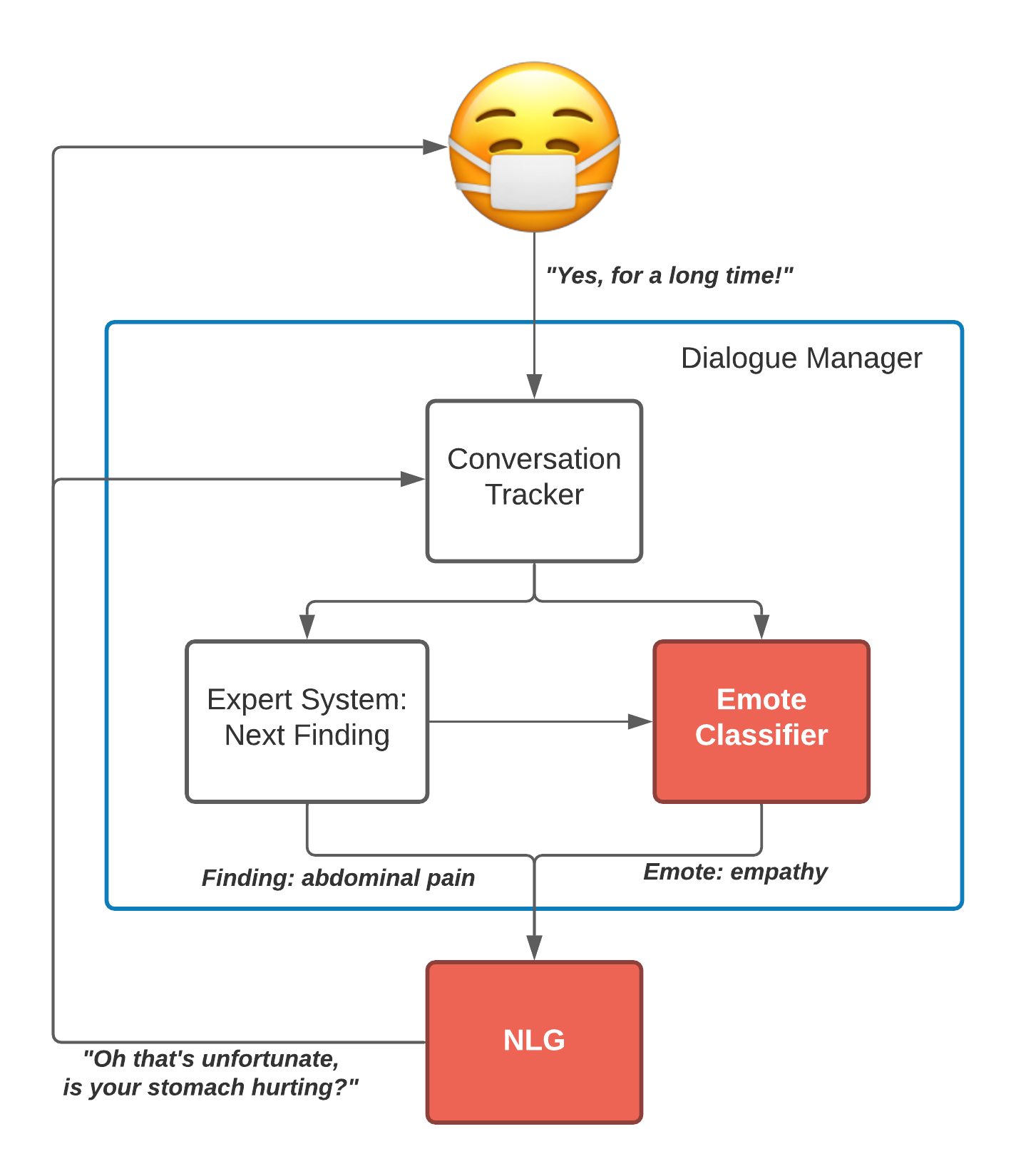}
    \vspace{-4mm}
    \caption{An overview of MEDCOD system presented in this paper, with main contribution areas highlighted in red.}
    \vspace{-7mm}
    \label{fig:model-overview}
\end{figure}
\vspace{-.1in}
We present details of \textbf{MEDCOD}, our medical dialogue system for history taking which combines expert-system-driven structured history taking (i.e. generate ``Next Finding" using a medical KB) with deep-learning-driven emotion classification and controllable NLG. This integration allows us to use the expert system to determine ``what" to ask (by the system to the user) in an explainable and auditable way, and to use the deep-learning components to determine ``how" to ask with human-like natural language. To enable this separation of ``what" and ``how" we developed a NLG module in the dialogue system that uses control codes provided by the expert system and the emotion understanding component to guide the formation of the NLG module's output.
\figureref{fig:model-overview} provides overview of  \textbf{MEDCOD}. Its \textbf{Dialogue manager}  consists of three components:
\vspace{-.1in}
\begin{enumerate}
    \item \textbf{Conversation tracker}: this component tracks patient demographic information, reason for encounter, what findings have been reported by the patient, the text of the previous questions, and the text of patient responses.
    \vspace{-.05in}
    \item \textbf{Next Finding}: this expert-system component takes patient demographics and patient findings and generates the target finding (\verb|next finding| control code) to be asked next by the NLG.
    \item \textbf{Emotion Classifier}: this model takes the conversation context and predicts the appropriate emote (\verb|emote| control code) to be used by the NLG model (\S~\ref{sec:model-emotion_classifier}). This component was trained using the Emote dataset  (\S~\ref{sec:emote_codes}).
\end{enumerate}
 
\paragraph*{NLG} component uses previous findings as well as control codes for the target finding and emote to generate a human-like NL question about the target finding (\S~\ref{sec:model-dialogpt}). This component was trained using the Medical Conversations dataset  (\S~\ref{sec:medical_conversation}).


\subsection{Emotion Classifier}
\label{sec:model-emotion_classifier}
During medical history taking, the patient may provide sensitive or emotionally charged information (e.g. severe pain); it is imperative that an automated dialogue system reacts and emotes appropriately to this information, similarly to how a human doctor would (\emph{e.g.} ``Oh that's unfortunate..."). When analyzing patient-provider medical conversations, we identified four broad classes of emote control codes that reflect emotional phrasing medical professionals use when talking with their patients. The control codes are \verb|Affirmative|, \verb|Empathy|, \verb|Apology|, and \verb|None| (see \S~\ref{sec:emote_codes} for more details). The goal of the emotion classifier is to predict the emote control code  based on the conversational context. The  conversational context contains three pieces of information: (1) previous question (2) patient response, and (3) target finding (which is the output of Next Finding module). 

The model consists of embedding the contexts independently (using a pretrained model) to capture the semantics of the entirety of text, and then learning a linear layer of predictors, on reduced dimensionality, over the emote control codes. We first independently embed the three pieces of context using Sentence-BERT (SBERT) \citep{reimers2019sentence} that takes as input a variable-size string (up to 128 tokens) and outputs a fixed-size vector containing semantically-meaningful feature values.  We then apply principal component analysis (PCA) \citep{pearson1901pca_orig, hotelling1933pca} to the embeddings of each input type independently and then concatenate the embeddings.  Finally, we train a logistic regression classifier over the four emote control code classes. 
The model is trained on the Emote dataset (\S~\ref{sec:emote_codes}). 


\subsection{Natural Language Generator}
\label{sec:model-dialogpt}
We developed the domain-specific NLG module of \textbf{MEDCOD} with three key constraints and goals:
\begin{enumerate}
    \item\textbf{Medical Consistency}: Generated questions by the system must ask only about the target finding (\emph{e.g.} if the target finding is ``abdominal pain" then an acceptable question would be ``Is your belly hurting?" while an unacceptable question would be ``Do you have severe abdominal pain?").  \vspace{-.1in}
    \item\textbf{Phrasing Diversity}: Generated questions must present phrasing variability, a major improvement over using templated questions (\emph{e.g.} if the target finding is ``abdominal pain",  instead of asking every time ``Is your belly hurting?", the model can generate alternative paraphrases such as ``Do you have pain in your belly?") 
     \vspace{-.1in}
    \item\textbf{Emotional Awareness}: Generated questions must be empathetic when appropriate. When gathering pertinent findings from the patient, we would like the NLG output to emote appropriately based on the context: did the patient say anything particularly difficult that we should empathize with (\emph{e.g.} a patient complaining about severe pain)? Are we about to ask a highly relevant (for a presenting symptom) but sensitive question (\emph{e.g.} checking with the patients if they have multiple sexual partners)?
\end{enumerate}

We achieve these three goals simultaneously by fine-tuning a pretrained DialoGPT model \citep{zhang2019dialogpt}. In the fine tuning process, we use control codes for dialogue prompts to help guide the NLG output \citep{Keskar2019} at inference-time. Apart from the control codes, we also prompt with the previous findings, patient's age and gender, as well as patient's reason for visit. The full control codes consist of the \verb|next finding| control code and the \verb|emote| control code. At training time, we use the Medical Conversations dataset (\S~\ref{sec:medical_conversation}). At inference time, the control codes are generated by the dialogue manager: \verb|next finding| control code comes from the Next Finding module while the \verb|emote| control code comes from the Emotion Classifier. 

\begin{figure*}[t]
\floatconts
  {fig:datasets}
  {\caption{Medical conversation and emote datasets construction.}}
  {%
    \subfigure[Schematic of the dataset construction.]{\label{fig:datasets-schematic}%
      \includegraphics[width=0.6\linewidth]{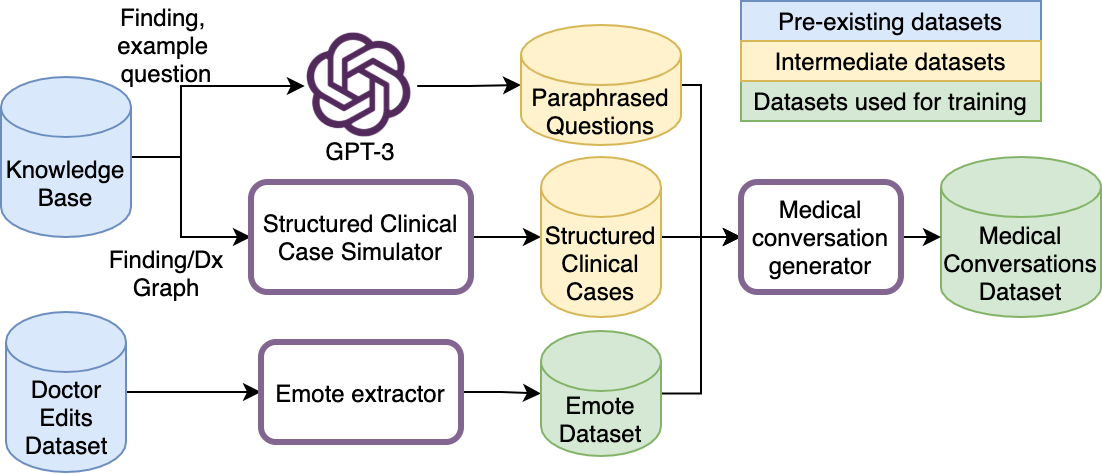}}%
    \qquad
    \subfigure[Examples illustrating training set construction. We append control codes during training to condition the models on them, which can then be used for guiding NLG during inference. ]{\label{fig:datasets-example}%
    \makebox[\linewidth][c]{
      \includegraphics[width=0.8\linewidth]{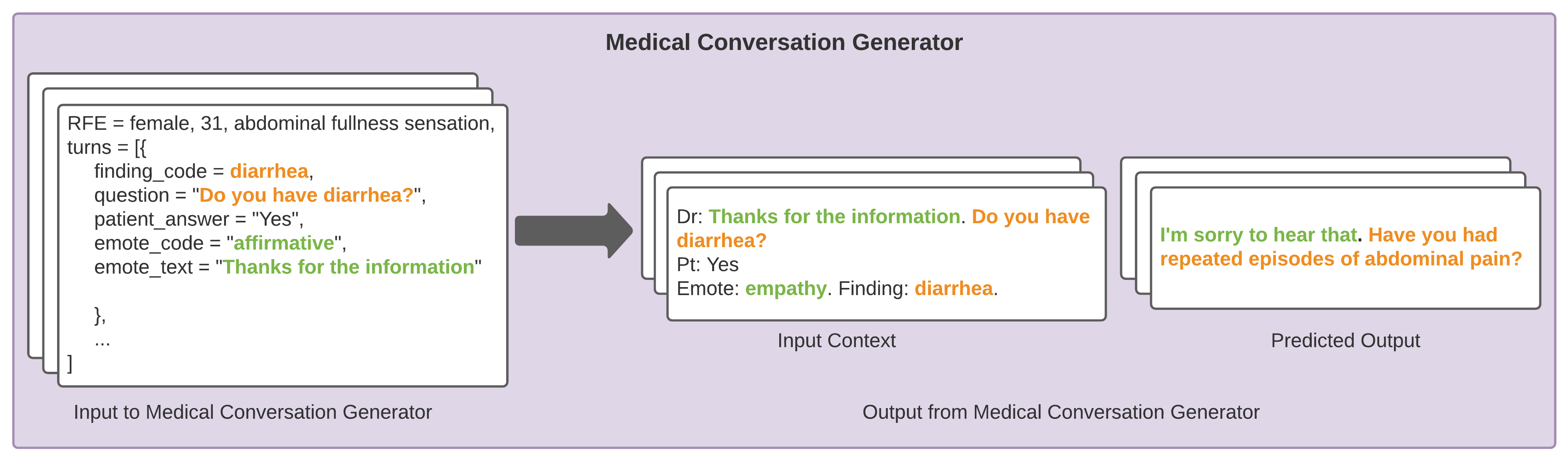}}
      }
  }    \vspace{-3mm}
\end{figure*}

\section{Datasets}
\label{sec:dataset}



The development of our medical dialogue system relies on a number of datasets. The process for constructing these datasets is presented in \figureref{fig:datasets-schematic}. We start with two preexisting datasets: KB and Doctor Edits dataset. The KB is used to generate two additional intermediate datasets, the Structured Clinical Cases dataset (\S~\ref{sec:struturded_data}) and the Paraphrased Questions dataset  (\S~\ref{sec:paraphrases}). The Doctor Edits dataset is used to construct the Emote dataset (\S~\ref{sec:emote_codes}). Finally the Structured Clinical Cases, Paraphrased Questions,  and Doctor Edits datasets are combined to produce the Medical Conversations dataset (\S~\ref{sec:medical_conversation}). The final datasets consist of the \textbf{Emote dataset} and \textbf{Medical Conversations dataset}. 
\vspace{-0.3cm}
\subsection{Pre-existing datasets}

\subsubsection{Knowledge Base} The knowledge base (KB) is an AI expert system similar to Quick Medical Reference (QMR)~({\it c.f.}~\cite{miller1990quick}) that is kept up-to-date by team of medical experts. Currently, it 
    contains 830 diseases, 2052 findings ($\mathcal{F}$)  (covering symptoms, signs, and demographic variables), and their relationships. It also contains human generated patient-answerable questions for ascertaining the presence for every finding. Finding-disease pairs are encoded as \emph{evoking strength (ES)} and \emph{term frequency (TF)}, with the former indicating the strength of association between the constituent finding-disease pair and the latter representing frequency of the finding in patients with the given disease. The inference algorithm of KB computes differential diagnosis and also facilitates the next finding to ask using \textbf{Next Finding} module.
\subsubsection{Doctor Edits dataset} This is a small in-house dataset containing 3600 instances of doctor-edited questions as well as doctor and patient dialogue turns preceding the doctor-edited questions. For training the emotion classifier, we perform a random 80/20 train/test split.

\subsection{Intermediate datasets}

\subsubsection{Structured Clinical Cases dataset} 
\label{sec:struturded_data}
Structured Clinical Cases dataset contains simulated cases that consist of patient demographic data (age, sex), the chief complaint also known as reason for encounter (RFE) and a set of findings with ``present" or ``absent" assertion. The data is produced using an in-house simulator, Structured Clinical Case Simulator (SCCS).  SCCS uses the KB as a clinical decision support system. 
Unlike previous works that simulate clinical cases for a desired diagnosis \citep{qmr-simulated-cases, Kao2018, ravuri18, Kannan2020}, SCCS starts with a finding and goes through the process of history taking to lead up to a diagnosis: It first samples demographic variables and a finding  $f \in \mathcal{F}$ that serves as the chief complaint/RFE. Then, it computes the differential diagnosis (distribution over the diseases given the finding) using the underlying expert system and then samples findings taking into account the differential diagnosis. The newly sampled finding is asserted randomly to be present ${\bf f}_{pos}$ or absent ${\bf f}_{neg}$ with a slight bias to absent. If asserted as present, then the findings that are impossible to co-occur are removed from consideration (\emph{e.g.} a person cannot have both productive and dry cough). The next iteration continues as before: computing differential diagnosis and then identifying next best finding to ask. The simulation for a case ends when a random number (5-20) of findings are sampled or the difference in score between the first and second ranked diagnosis is at least 20 (a desired minimum margin under expert system). Simulated samples with margin higher than that are added to the Structured Clinical Cases dataset. See \S~\ref{apd:patient_case} for an example simulated clinical case.
\vspace{-.15in}
\subsubsection{Paraphrased Questions dataset}
\label{sec:paraphrases}
The Paraphrased Questions dataset contains findings and an associated set of questions, these questions being different ways (paraphrases) to ask about the finding (for examples see Table~\ref{tab:gpt3_paraphrases}). The goal of this dataset is to imbue variability into the NLG model with examples of different question phrasings.


We use carefully primed GPT-3 \citep{brown2020} to generate a large number of candidate questions for each finding. We curate a small but \emph{diverse} set of thirty findings and manually paraphrase the expert-written question already available from the KB.
We randomly sample 10 findings for priming \citep{liu2021, chintagunta2021} GPT-3 to paraphrase new unseen findings. See Appendix \ref{apd:sec-gpt3_invocations} for example invocations. To restrain the generations but still acquire a diverse set of paraphrases, we limited the output to a single paraphrase at a time. We repeatedly invoke GPT-3 (\verb|temp=0.65|) until we have the desired number of distinct paraphrases, each time priming GPT-3 with random sample of ten findings.

The key strength of this approach is the minimal human effort required; only a single manually-written paraphrase is required for each finding in our small set, which is then used as guidance for GPT-3 to mimic the task on new findings.  We manually validate the candidate questions using in-house \doctor s by asking them to label if the candidate question is medically consistent with the target finding, and keep only those that are. We achieved 78\% correctness of finding-question pairs. Analysing the failure cases, we found that the error was either due to minor grammar issues or bad timing (i.e., "Are you sleepy?", which implies right now as opposed to intermittently throughout the day). We collected question paraphrases for the 500 most frequent findings in the Structured Clinical Cases dataset.

\subsection{Final datasets for training}
\subsubsection{Emote dataset}
\label{sec:emote_codes}
The Emote dataset contains a set of emote phrases, their corresponding emote control codes, and patient and \doctor~dialogue turns that preceded the use of the emote phrase. This dataset is directly used to train the Emotion Classifier (\S~\ref{sec:model-emotion_classifier}). The emote phrases are directly extracted from \doctor~messages, while the \verb|emote| control codes are manually assigned to each emote phrase. 

We mined the Doctors Edits dataset for \doctor~messages that express emotion and identified three broad clases of \verb|emote| control codes: 
\begin{itemize}
    \item \textbf{Affirmative:} A neutral confirmation of the user's response, flexible to be used in many conversation situations (e.g. thanks for the input). 
    \vspace{-.25in}
    \item \textbf{Empathy:} A more emotionally charged response, implying something negative/painful about the conversation. 
    \vspace{-.1in}
    \item \textbf{Apology:} This is an apology for asking a personal or sensitive question (e.g., "Sorry for asking a personal question, do you have multiple sexual partners?"). 
\end{itemize}
We also included  \textbf{None} as an emotion code to reflect no emotion is added. We associated with each emote code a set of emote phrases that are frequently used by \doctor s to express these codes. We provide additional details on data mining for emotion from conversations and examples of emote phrases corresponding to codes in  Appendix~\ref{apd:emote_dataset_construction}.

\subsubsection{Medical Conversations dataset}
\label{sec:medical_conversation}
The Medical Conversations dataset consists of dialogue context, \verb|next finding|~and ~\verb|emote| control codes, and medical finding questions with emotional responses; the NLG model is trained on this dataset.
 
\figureref{fig:datasets-example} provides a walk through for constructing Medical Conversation samples. To generate dialogue context we sample from the Clinical Cases dataset. From this we extract the RFE, patient demographic information, target finding, and the structured preceding finding and patient response. The preceding finding and the target finding are converted into a preceding \doctor~question and target question in the following manner. We sample from the Paraphrased Questions dataset a question corresponding to the finding. We then randomly choose an emote control code and a random corresponding emote phrase from the Emote dataset. The emote phrase is then prepended to the finding-based-question. For the patient response we simply select “Yes” or “No”, based on the assertion attached.

\section{Experimental Results}
\label{sec:experimental_results}
In this section we present both subjective and objective evaluation results which robustly demonstrate the improved output from \textbf{MEDCOD} when compared to counterparts that use the fixed-template approach to asking questions or can not emote.

\subsection{Experimental Setup}

\subsubsection{Training details}
\noindent \textbf{NLG}: We use a pretrained \verb|DialoGPT-Medium| from HuggingFace \citep{huggingface2021} 
as our underlying NLG model. We train on 143,600 conversational instances, where each instance has only one previous conversation turn as context. We use a batch size of 16 with 16 gradient accumulation steps for an effective batch size of 64, for 3 epochs with a learning rate of 1e-4 and ADAM optimiser. 

\noindent \textbf{Emotion Classifier}: We apply  pretrained \verb|paraphrase-mpnet-base-v2| SBERT for embedding the conversational contexts. The Logistic Regression model is trained with \verb|C=10| and class re-weighting (to compensate for the data skew of the training data \S~\ref{sec:emote_codes}). PCA is applied down to 70 components.

\subsubsection{Ablations of MEDCOD}
We ablate MEDCOD by varying data/control codes supplied to each underlying NLG model during training, with all other parameters kept consistent. This allows us to understand the importance of variability, medical consistency and ability to emote. We use \textbf{Expert} to denote the variant of MEDCOD in which the NLG module is trained only on expert questions (single question per finding). \textbf{MEDCOD-no-Emote}'s NLG module is trained on the Medical Conversations dataset (\S~\ref{sec:medical_conversation}) with paraphrases but no emote codes while \textbf{MEDCOD} is our feature-complete dialog system trained on the entire Medical Conversations dataset including emote.




\begin{table}
    \centering
    \begin{tabular}{llll}
    \toprule
                    & A     & B     & Equal  \\
                    \midrule
Total Pts         & \textbf{63}          & 30          & -           \\
Mut. Excl. Pts  & \textbf{49}  & 16  & 25  \\
 & (54.4\%) & (17.8\%) & (27.8\%) \\
\midrule
\multicolumn{3}{l}{\textit{Aggregated with Majority Voting Applied}}     &             \\
Total Pts    & \textbf{24}          & 6           & -           \\
Mut. Excl. Pts. & \textbf{20} & 2 & 8 \\
 & (66.7\%) & (6.6\%)   & (26.7\%) \\
\bottomrule
\end{tabular}
    \vspace{-4mm}
    \caption{End2End Evaluation comparing between \textbf{MEDCOD} (A) and \textbf{Expert} (B)}
        \vspace{-8mm}
    \label{tab:results-holistic_comparison}
\end{table}

\subsection{End2End Evaluation: Main Results}
\label{sub:results-end2end_comparison}
This is our main evaluation that is targeted at understanding if the patient experience on the end-to-end system can be improved by providing them with a more natural conversational dialog. For this, we instantiate two identical medical dialog interfaces with different driving systems: \textbf{Expert} and \textbf{MEDCOD}. A set of 30 commonly occurring chief complaints along with demographic information such as age and gender were collected from a telehealth platform. We recruited five \doctor s for the labeling task such that each \doctor~will go through the conversational flow for 18 chief complaints, giving three labels for each case. 

\noindent \textbf{Labeling instruction}: While the focus is on patient experience, we engaged \doctor s because of the dual patient/doctor role-play for this evaluation. When they start on a chief complaint, they were to choose a relevant final diagnosis and answer questions to substantiate that final diagnosis; this ensures that the sequence of questions asked during conversation are clinically grounded. While doing so, they were also acting as a patient, answering and responding (e.g., by volunteering extra information) as someone presenting with the symptoms would. The \doctor s converse simultaneously with the \textbf{Expert} and \textbf{MEDCOD} systems (the UI presents an anonymized A/B label) by providing identical answers for each conversation step between the two interfaces (but different answers between steps).

Once they perform 10 question responses or the conversation terminates (due to reaching a diagnosis), the encounter is over and they label each instance as follows: \textit{Enter a 1 for either Model A or Model B, based on how you think a new patient using the service for an ailment would feel. If you prefer the encounter with Model A, enter a 1 in the Model A column and 0 in the Model B column, and vice versa if you prefer Model B.} To avoid spurious ratings when the two models are very similar, we also allowed the same grade to be given to both models if they were equally good/bad, but required a comment explaining the decision.

\noindent \textbf{Results}: Table \ref{tab:results-holistic_comparison} shows the evaluation results (refer to Appendix \ref{apd:conv_comparison} for a full conversation example).  When simply summing up the scores, \textbf{MEDCOD} achieves a score of 63 (max 90) --- over twice as high as \textbf{Expert}. When separating scores into instances where one model is picked exclusively over the other or both are rated equally (``Mut. Excl. Pts"), we see a similarly strong result for our model; in over half the conversations enacted, \textbf{MEDCOD} is preferred holistically over \textbf{Expert}, while only 17.8\% of the time is \textbf{Expert} preferred. When we inspect the difference, its often the case that \textbf{MEDCOD} emoted with \verb|Affirmative| when not emoting (\verb|None|) would have been more appropriate. 

We also considered majority voting for each of the patient complaints, which shows an even more exaggerated improvement by our model. Two-thirds of the time (66.7\%), \textbf{MEDCOD}  is exclusively preferred over \textbf{Expert}, while only 6.6\% of the time latter is preferred. In roughly one-fourth of the chief complaints, both models are rated equally.

In the majority of cases, \textbf{MEDCOD} is preferred, indicating that within an automated dialog situation, the contributions discussed in this paper provide a marked improvement to patient experience. 


\subsection{NLG Module Evaluation}
\label{sec:nlg_eval}
The goal is to evaluate MEDCOD and its ablations individually along three important aspects for medical dialog (c.f. \citep{rashkin2018empathy}): 
\begin{enumerate}
    \item \textbf{Medical Consistency: }How well does the question capture the clincial implication of the target finding?
    \item \textbf{Fluency:} How \textit{fluent/grammatically correct} is the candidate question?
    \item \textbf{Empathy:} How \textit{empathetic/emotionally appropriate} is the candidate question, given the conversational context and the finding to ask next?
\end{enumerate}

To collect the data for this evaluation, we begin with an in-house dataset of conversations from a tele-health platform. We decompose each conversation into three-turn instances (same form as Emote Dataset \S\ref{sec:emote_codes}), then attach an emote control code to each instance by performing prediction with the Emotion Classifier. To exaggerate the difference between instances, we only keep instances where the predicted class' probability $>0.8$. We then randomly sample 25 instances from each of the four predicted classes to create our final set of 100 evaluation instances. 
Finally, we generate a candidate question for the instances by passing the conversation context to each of the model variants for generation. A team of five \doctor s label each example along each of three axes on a scale of 1 to 5.

\begin{table}
\makebox[\linewidth][c]{
    \begin{tabular}{l l l l}
    \toprule
    Model  & Medical & Fluency & Empathy \\
   Variant& Accuracy & & \\
    \midrule
    \textbf{Expert} & 4.956 & 4.942 & 2.772 \\
    \midrule
    \begin{tabular}[c]{@{}l@{}}\textbf{MEDCOD-no}\\ ~~~\textbf{-Emote} \end{tabular} & $4.882^*$ & $4.832^*$ & 2.806 \\
    \textbf{MEDCOD} & $4.872^*$ & $4.730^*$ & \textbf{3.892}$^*$ \\
    \bottomrule
    \end{tabular}
    }
    \vspace{-4mm}
    \caption{NLG evaluation - \textbf{MEDCOD} shows significant improvement in empathy and added variability without sacrificing other aspects}
    \vspace{-10mm}
    \label{tab:results-individual_gen}
\end{table}
\noindent {\bf Results}: Table \ref{tab:results-individual_gen} provides the comparative results. \textbf{MEDCOD} scored significantly higher in Empathy, showing that  the  Emote  dataset  additions  improve  human-evaluated  empathy  in  a significant  way. This result also indicates that the \verb|emote| code is appropriately predicted by the Emotion Classifier.

There  are  many  correct  ways  to  query  a  finding, however the \textbf{Expert} model is trained on data with precisely  one  way,  which  is  expert-annotated,  so is likely to have optimal medical consistency (and also be the most fluent for the same reason).  Because of this, we view \textbf{Expert} as close to best performance achievable  along  Medical  Accuracy  and  Fluency. \textbf{MEDCOD} and \textbf{MEDCOD-no-Emote} are still comparable to \textbf{Expert} indicating that the variations in how questions are framed do not significantly affect medical accuracy or fluency.  As expected, given that its impossible  to  encode  empathy  preemptively  (in the expert-annotated or paraphrased questions), \textbf{Expert} and \textbf{MEDCOD-no-Emote} score low on empathy.  Note that it is not always necessary to emote, hence they receive non-zero score.

Appendix ~\ref{tab:qual_comparison} provides qualitative examples of comparing generations from model variants where  \textbf{MEDCOD} is expressive and incorporates empathy.

\subsection{Emotion Classifier Evaluation}

Evaluating emotion is difficult as it is subjective and can be multi-label: in a situation, there may be multiple ``correct" ways to emote so comparing predictions to a single ground-truth label (i.e., physician's emote) is unlikely to give an accurate notion of performance. We instead measure the \textit{emotional appropriateness} using a small team of \doctor s.

\begin{figure}
    \centering
    \includegraphics[width=0.8\linewidth]{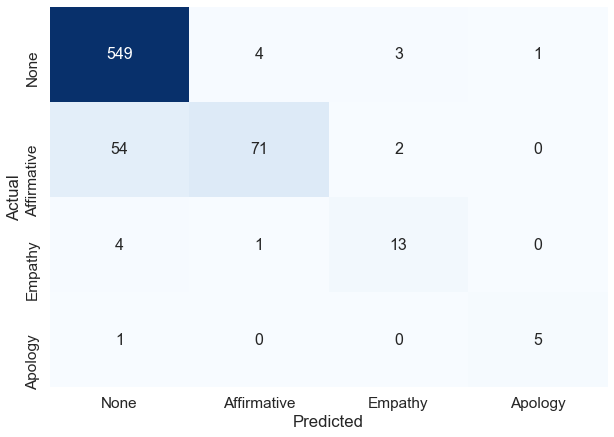}
    \vspace{-4mm}
    \caption{Emotion Classifier Evaluation}
    \vspace{-5mm}
    \label{fig:confusion_matrix}
\end{figure}

For each instance in the Emote dataset test split, we pass the predicted \verb|emote| control codes to a team of three \doctor s. They are tasked with labelling whether the \verb|emote| code is appropriate, given the previous context in a conversation (input to our model). When the emote is not appropriate, an alternate emote is suggested by the labeller. We use majority voting on this data to obtain the final label, creating an alternate \textit{human-augmented} test set. We evaluate the model's predictions against this \textit{human-augmented} test set; Figure \ref{fig:confusion_matrix} shows the confusion matrix on the full Emote dataset test split and Table~\ref{apd:tab-emote_clf_report} in Appendix shows the complete results.

On the \textit{human-augmented} test set, our Emotion Classifier reached \textbf{0.9} accuracy with macro-F1: \textbf{0.8} and PR-AUC: \textbf{0.69}. Looking at precision/recall statistics, each non-\textit{None} emote class (and the model predictions overall) achieved precision $>=$ recall, which is desired due to the high cost of inappropriate emoting; we want high confidence when we actually emote something, otherwise we should safely emote \textit{None}. It should be noted that we did not tune the classification boundary but simply took the max-probability class as prediction; a high prediction threshold (e.g., 0.8) would further increase precision.

The confusion matrix (Figure \ref{fig:confusion_matrix}) illustrates a similarly strong picture of our Emotion Classifier's performance; the large majority of non-\textit{None} predictions are correct. There are 54 `incorrect' \textit{None} predictions by our Emotion Classifier, however, these are low- or zero-cost mispredictions, as it is always appropriate not to emote (the same cannot be said for \textit{Empathy}, for example, where it can be very costly to empathise when not appropriate).







We also analyzed how conversational context affects predicted emotion. See Appendix~\ref{sec:emote-attribution} for complete attribution details. We find that \verb|empathy| is strongly influenced by the previous patient response, \verb|apology| by the next question and \verb|affirmative| by all three parts of the input.  



\section{Conclusion \& Future Work}
\label{sec:conclusion}
We introduced \textbf{MEDCOD}\footnote{\url{https://github.com/curai/curai-research/tree/main/MEDCOD}}, a novel approach to developing medical dialog systems, which combines the traditional modular design with a deep learning based NLG model. In particular, our approach allows us to incorporate, in a medically consistent fashion, the knowledge of medical findings and an appropriate emotional tone when generating human-like NL expressions through the use of their respective control codes, both provided by the dialog manager. The highly positive experimental results presented demonstrate the effectiveness of our approach.

Our current approach still leaves a number of open problems. A broad problem for making the dialogue even more human-like is the effective use of implied references; as an example, when a patient mentions they have diarrhea, the system would ask them ``Is it bloody?" as a follow-up question, as opposed to ``Do you have bloody diarrhea?". Alternatively, historical context references can be used: asking ``You mentioned you have headache. Is this recurrent?" instead of ``Do you have a recurrent headache?".  

Another area of improvement in our current system is increasing the granularity of emotion classes. This may include both the addition of new classes and incorporating a notion of emotional intensity, e.g., ``OK'' has a different strength than ``Thank you!''

In the longer term, our dialogue system should accommodate speech recognition and speech generation. This presents a new class of challenges such as emotion detection from speech and synthesis of properly empathetic speech. Multimodal inputs beyond text and speech may further advance the functionality of dialogue systems, especially in the user-interface aspect \citep{Yu2008}. For example, the video-feed of a patient may present additional opportunities for refining our dialogue manager module by collecting non-verbal cues from the patient.

\bibliography{main}

\counterwithin{figure}{section}
\counterwithin{table}{section}
\clearpage
\appendix

\section{Simulated Clinical Case}\label{apd:patient_case}
\begin{table}[h]
\begin{lstlisting}
{
    "id": 0,
    "age": [
      "young adult (18 to 40 yrs)"
    ],
    "gender": [
      "male"
    ],
    "RFE":[ "abdominal fullness sensation+"]
    "findings": [
      "diarrhea, chronic+",
      "abdominal pain, recurrent attacks+",
      "lactose intolerance-",
      "gluten intolerance-",
      "marijuana use+",
      "relieved by hot shower+",
      "worse in the morning-",
      "vomiting, recurrent+",
      "intermittent+",
      "anxiety+",
      "sweating increase+",
      "thirst increase+",
      "weight loss+",
      "epigastric abdominal pain-",
      "chronic (> 4 weeks)+"
    ]
}
\end{lstlisting}
\label{tbl:patient_case}
\caption{A clinical case simulated from KB. }
\end{table}
\section{Question Paraphrasing using GPT-3}\label{apd:gpt3_paraphrases}

\begin{table}[H]
    \centering
    \begin{tabular}{l l}
    \toprule
        Finding & GPT-3 generated questions \\
        \midrule
        anxiety & Are you anxious? \\
        & Do you have anxiety? \\
        & Have you been experiencing \\ & ~any anxiety? \\
        & Are you feeling nervous or \\& ~anxious? \\
        \midrule
        back pain & Do you feel pain in your back? \\
        & Is your back hurting?	 \\
         & Does your back hurt? \\
         & Are you experiencing pain \\ & ~in your back? \\
         \bottomrule
    \end{tabular}
    \caption{Examples of question paraphrases generated by GPT-3. }
        \vspace{-8mm}
    \label{tab:gpt3_paraphrases}
\end{table}

\section{Details of Emote dataset construction}\label{apd:emote_dataset_construction}
We generated the emote dataset using an in-house Doctors Edits dataset, which contains 3600 instances of \doctor-edited questions and their preceding \doctor~and patient messages. \doctor-edited questions are templated questions from KB and then were subsequently edited by the professionals based on the context of the conversation. 

These edits are typically done to impart additional emotion to the text (emote phrase), although some of the edits are made for more pragmatic reasons (e.g. improve question readability). To extract the emote phrase, we use a simple heuristic method based on the assumption that the edited question is of the form: \verb|[emote phrase] KB question| \verb|[additional information]|, \emph{e.g.} \textit{"Oh I'm sorry to hear that. Do you have flushing? That is, do your arms feel warmer than usual?}. We simply split the edited question on punctuation and identify which part of the split question is closest to the KB question by fuzzy string matching\footnote{\url{https://github.com/seatgeek/fuzzywuzzy}}; once the most similar question section is identified, the emote phrase is returned as the preceding sub-string to this section. (Algorithm \ref{alg:model-emotion_extraction} in Appendix for details). The accuracy of this simple algorithm for our task was evaluated manually and shown to achieve 99.4\% accuracy within our limited domain of conversation.  Table~\ref{tab:emote_paraphrases} presents example emote language phrases corresponding to the emote control codes. 

The dataset has a class imbalance towards \textbf{None}, indicating it is often not necessary to emote, and when one is emoting, \textbf{Affirmative} is the most common.
\begin{algorithm}[t]
    \caption{Emotional Addition Extraction}
    \label{alg:model-emotion_extraction}
    \begin{algorithmic}[1]
        \renewcommand{\algorithmicrequire}{\textbf{Input:}}
        \renewcommand{\algorithmicensure}{\textbf{Output:}}
        \REQUIRE Default question $Q_d$, Edited question $Q_e$
        \STATE Initialize $splits =$ split $Q_e$ on punctuation
        \STATE Initialize $scores =$ empty array
        \FOR{$s \in splits$}
            \STATE $scores[s] \gets \textsc{FuzzyMatchScore}(s, Q_d)$
        \ENDFOR
        \STATE $i_q \gets argmax(scores)$
        \RETURN $splits[:i_q]$ \COMMENT{Return everything preceding the (most likely) question}
    \end{algorithmic} 
\end{algorithm}

\begin{table}
    \centering
    \begin{tabular}{l l}
    \toprule
        Control code & Emote language\\
        \midrule
        affirmative & Thanks for the input \\
        & Okay \\
        & I see \\
        & Got it \\
        \midrule
    empathy & Sorry about that \\
        & That's concerning \\
         & Okay, I'm sorry to hear \\
         \midrule
         apology & I am sorry for asking \\
          & I apologise if this is personal \\
          & I am sorry for asking if it sounds \\ & ~personal but may I know \\
         \bottomrule
    \end{tabular}
    \caption{Examples of emote codes and example sentences that are mined from real world medical conversations. See text for details}
    \label{tab:emote_paraphrases}
\end{table}

\section{Emotion Classifier Performance}

\begin{table}[H]
\makebox[\linewidth][c]{
\begin{tabular}{lllll}
\toprule
 & Precision & Recall & F1 & Support \\
 \midrule
\textit{None} & 0.90 & 0.99 & 0.94 & 557 \\
\textit{Affirmative} & 0.93 & 0.56 & 0.70 & 127 \\
\textit{Empathy} & 0.72 & 0.72 & 0.72 & 18 \\
\textit{Apology} & 0.83 & 0.83 & 0.83 & 6 \\
\midrule
Accuracy & - & - & 0.90 & 708 \\
Macro Avg & 0.85 & 0.78 & 0.80 & 708 \\
Weighted Avg & 0.90 & 0.90 & 0.89 & 708 \\
\bottomrule
\end{tabular}
}
\label{apd:tab-emote_clf_report}
\caption{Classification scores for Emotion Classifier on human-augmented Emote dataset.}
\end{table}

\section{Dissecting Emotion Classifier: Emotion Attribution}
\label{sec:emote-attribution}

As we mentioned in \S~\ref{sec:model-emotion_classifier}, we used logistic regression as the final classifer for the emote classifier, where: 
\begin{equation}
\text{logits}(p_{i}) =  \sum_{j} M_{ij}^{T} {\bf x}_{j} + b_{i},
\label{eq:logits}
\end{equation}
where $p_{i}$ is the probability of the $i^\text{th}$ control code and  index $j$ represents the input source (previous question, previous patient response, target finding), $M_{ij}$ is the learned coefficient vector corresponding to $i^{th}$ class and $j^{th}$ source, ${\bf x}_{j}$ is SBERT embedding vector corresponding to $j^{th}$ input source, after dimensionality reduced by PCA, and $b$ are learned biases. 

We analyzed how conversational context affects predicted emotion. Using eq.~\ref{eq:logits}, we  compute the contribution of each input source $j$ for each output control code $i$ by looking at the individual summands $M_{ij}^{T} {\bf x}_{j}$. The biases for each of the four classes are the following: None = 3.27,  Affirmative = 0.88, Empathy = -1.66 , Apology = -2.49. 

\begin{figure}[h!]
    \centering
    \includegraphics[width=0.4\textwidth]{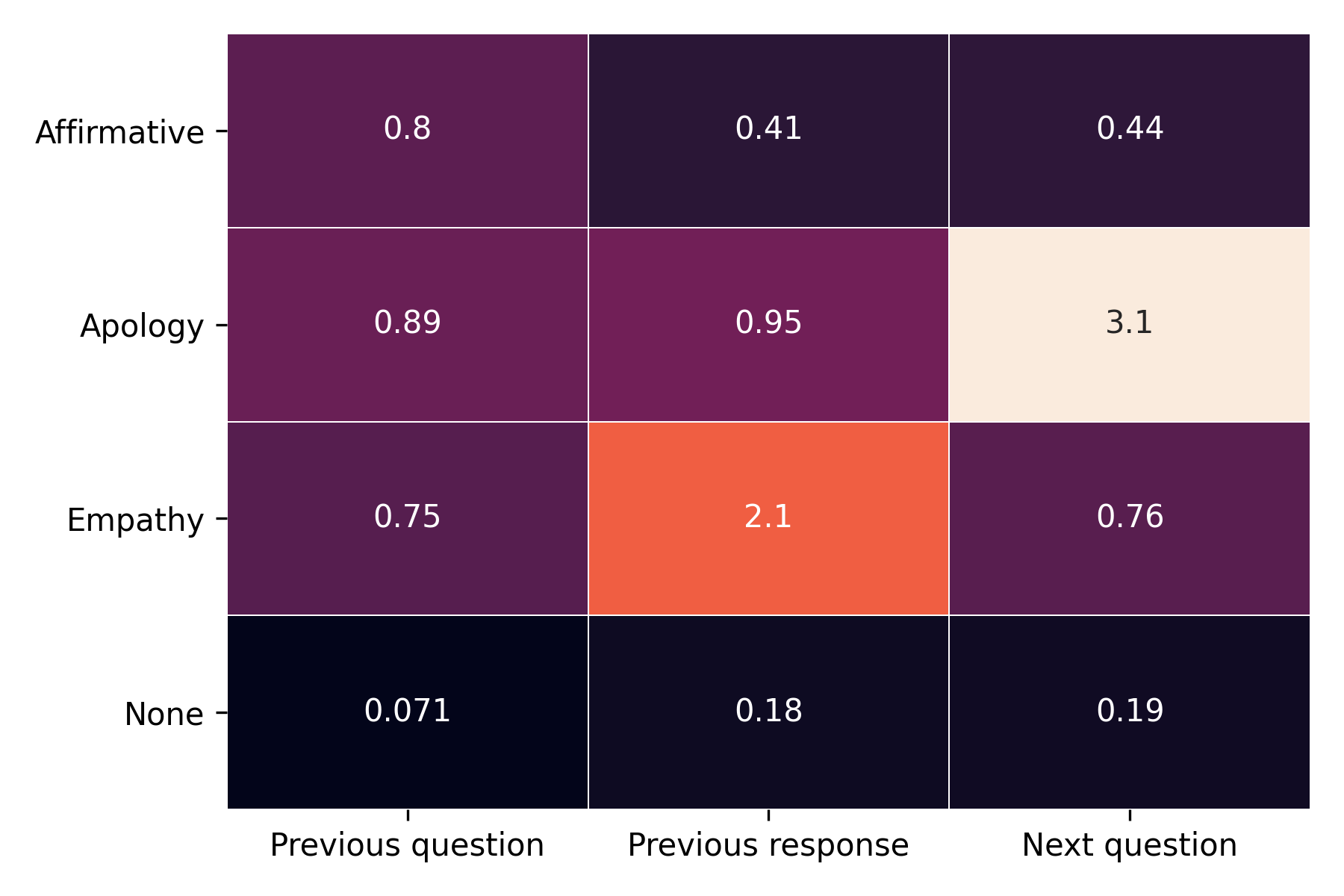}
    \caption{Contribution of input source to emote classification.}
    \label{fig:winning-contrib}
\end{figure}

\figureref{fig:winning-contrib} shows the mean contribution of each input source to the predicted output control code ($i^*= \argmax_i{{\bf p}}$). We find that  the apology and empathy  are strongly predetermined by a single input source. For the apology class main contributor is the next question, this is consistent with expectation ({\it i.e} we apologize for asking personal and embarrassing questions). For the empathy class the main contributor is the previous response, which again matches intuition that we show empathy if the patient response involves significant negative sentiment. For affirmative class, all three types of inputs are taken into account, with previous question has the largest contribution. Intuitively, this makes sense that we may prioritize responding to the previous answer based on not only what the exchange has been so far, but also what we are going to ask next. 
For the none class, none of the input sources contribute significantly, this is consistent with the none class being driven by the bias (3.27!) -- when the model doesn't predict an emote it is because none of the input sources contributed to an alternative class.
\clearpage

\onecolumn
\section{Generalization of MEDCOD's NLG Component}\label{apd:qual_results}
One of the key benefits provided by a neural language generator like DialoGPT over a simple lookup-based system is its ability to generalize to new unseen findings; the broad world knowledge learned through pretraining gives the model an ability to generate appropriate questions for new findings. 
Fig.~\ref{apd:tab-extrapolation} provides qualitative results on our NLG model for various `types' of input findings. 

\begin{table*}[h]
\begin{tabular}{lll}
\toprule
Category & Finding & Generation \\
\midrule
\multirow{4}{*}{\begin{tabular}[c]{@{}l@{}}Unseen semantically\\ meaningful findings\end{tabular}} & knuckle pain & Do you have any pain in your knuckles? \\
 & spider bite & Have you been bitten by a spider recently? \\
 & fell off motorbike & Have you recently fallen off a motorbike? \\
 & broken arm & Is your arm broken? \\
 \midrule
\multirow{3}{*}{Unseen fake finding} & \multirow{2}{*}{dysparatosis} & Do you suffer from dysparatosis \\
 &  & Have you been diagnosed with dysparatosis? \\
 & dysparatosis (temp=0.9) & Do you have difficulty having a bowel movement? \\
 \midrule
\multirow{4}{*}{\begin{tabular}[c]{@{}l@{}}Misc. findings \\ showcasing\\ general world \&\\ grammar knowledge\end{tabular}} & eaten too much cheese & Have you recently eaten too much cheese? \\
 & lost my phone & Have you lost your phone? \\
 & slept through my alarm & Did you miss your alarm? \\
 & \begin{tabular}[c]{@{}l@{}}slept through my alarm\\ (temp = 0.9)\end{tabular} & Have you been having trouble sleeping? \\
 \bottomrule
\end{tabular}
\label{apd:tab-extrapolation}
\caption{Examples of \textbf{MEDCOD} generating presence questions for unseen findings (temperature=0.6 unless specified). The model is able to generate a question medically consistent with the input finding, despite never seeing it during training, showing the NLG model's ability to generalize. When generating for a finding which it can't get semantic information from (e.g., a fake disease name), it defaults back to generic "Do you have [disease]" style questions, unless we use a higher temperature in which case we get unexpected results. The model can also generate questions for non-medical findings (e.g., \textit{lost my phone}); in these cases it again can generate a semantically consistent question with the "finding", although generations lose accuracy when temperature is raised.}
\end{table*}

\clearpage
\onecolumn
\section{Qualitative comparison of generated questions by three MEDCOD Instantiations}

\begin{table*}[!ht]
\makebox[\textwidth]{

\begin{tabular}{llll}
\toprule
Model                          & Finding                               & Emote                        & Generated Responses (multiple invocations)                               \\
\midrule
\multirow{4}{*}{\textbf{Expert}}        & \multirow{2}{*}{recurrent headache}   & \multirow{2}{*}{affirmative} & \textit{Do you have headaches that come and go often?}                    \\
                              &                                       &                              & \textit{Do you have headaches that come and go often?}                    \\
                              & \multirow{2}{*}{generalized weakness} & \multirow{2}{*}{empathy}     & \textit{Does your entire body feel weak?}                                 \\
                              &                                       &                              & \textit{Does your entire body feel weak? }                                \\
                              \midrule
\multirow{4}{*}{\textbf{MEDCOD-no-Emote}} & \multirow{2}{*}{recurrent headache}   & \multirow{2}{*}{affirmative} & \textit{Do you have headaches that occur on a regular basis?}             \\
                              &                                       &                              & \textit{Are your headaches recurring?}                                    \\
                              & \multirow{2}{*}{generalized weakness} & \multirow{2}{*}{empathy}     & \textit{Are you feeling weak in your entire body?}                        \\
                              &                                       &                              & \textit{Is your entire body weak?}                                        \\
                              \midrule
\multirow{4}{*}{\textbf{MEDCOD}}          & \multirow{2}{*}{recurrent headache}   & \multirow{2}{*}{affirmative} & \textit{Thanks for the input. Does your headache keep coming back?}       \\
                              &                                       &                              & \textit{Okay. Is your headache recurrent?}                                \\
                              & \multirow{2}{*}{generalized weakness} & \multirow{2}{*}{empathy}     & \textit{Sorry to know that. Are you weak all over?}                       \\
                              &                                       &                              & \textit{That's worrisome. Do you experience weakness all over your body?}
\end{tabular}
}
\label{tab:qual_comparison}
\caption{Qualitative comparison of generated questions by the three NLG models. Note the lack of diversity in the \textbf{Expert} model and lack of emotion in the \textbf{MEDCOD-no-Emote} model. The questions generated by \textbf{MEDCOD} show the greatest combination of emotion and variation. For brevity, not shown is the previous turn of the conversation that is also used as input.}
\end{table*}
\clearpage

\section{Full Conversation Comparison}
\begin{figure*}[h]
    \centering
    \includegraphics[width=0.8\linewidth]{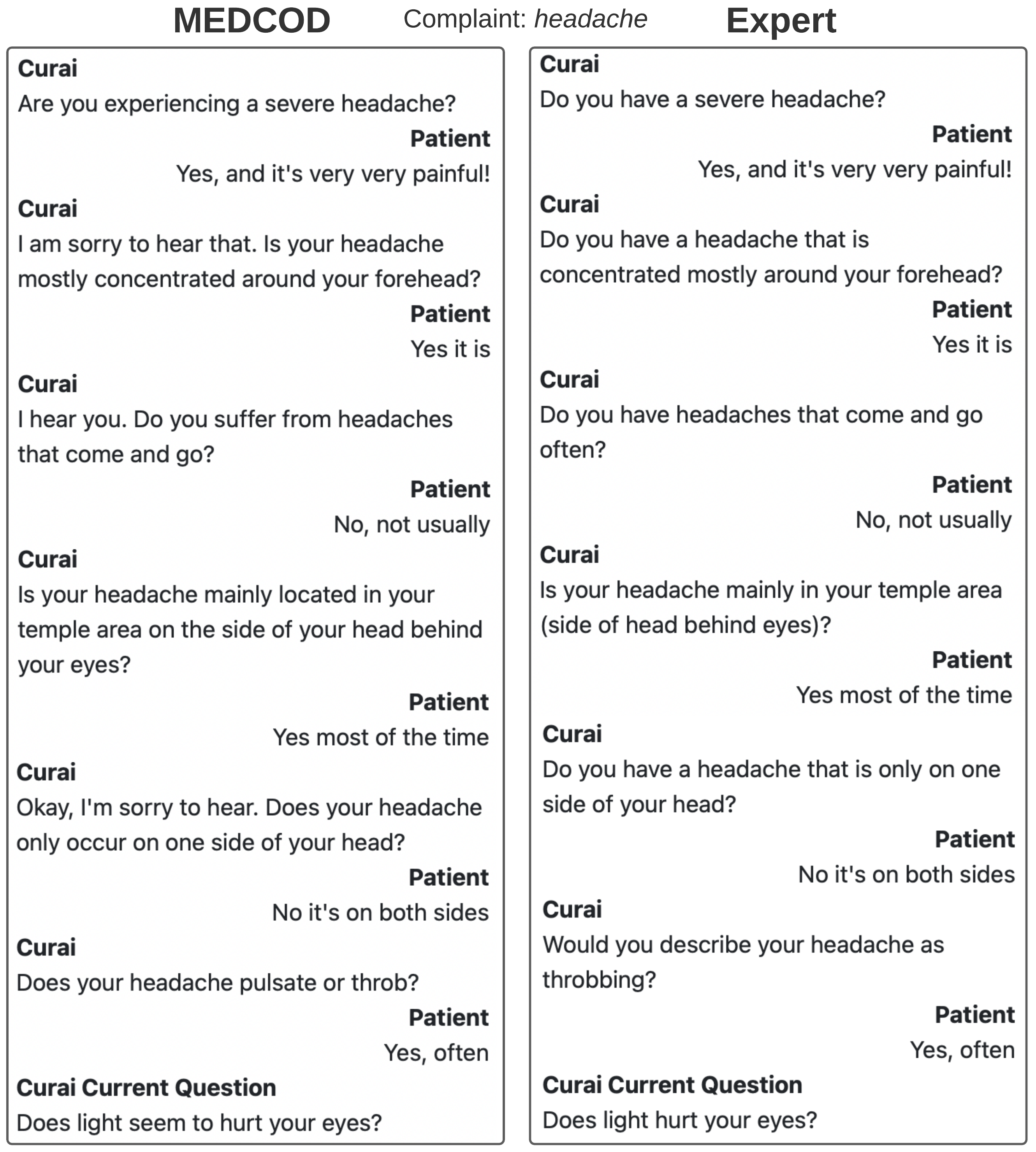}
    \caption{\textbf{MEDCOD} and \textbf{Expert} system used for mock patient case. Note the more natural conversation flow with \textbf{MEDCOD} due to its emotive responses and variation in question framing.}
    \label{apd:conv_comparison}
\end{figure*}

\clearpage
\section{GPT-3 Invocations}
\label{apd:sec-gpt3_invocations}

Following are some prompts given to GPT-3 to generate new paraphrases for the \textbf{progressive paralysis} finding. Bold is the generated response by GPT-3. These are used to generate the following five distinct paraphrases:
\begin{itemize}
    \item Are you noticing a progressive weakness on one side of your body?
    \item Have you noticed a gradual weakening of one side of your body?
    \item Have you noticed a weakness on one side of your body that seems to be getting worse?
    \item Has your weakness on one side of your body been getting progressively worse?
    \item Is one side of your body becoming increasingly weaker than the other?
\end{itemize}

\begin{figure*}[!ht]
\floatconts
  {apd:fig-gpt3_prompts}
  {\caption{Individual GPT-3 Prompts for the same finding, showcasing the prompt engineering and diversity of output.}}
  {%
    \subfigure[First invocation]{
      \includegraphics[width=\linewidth]{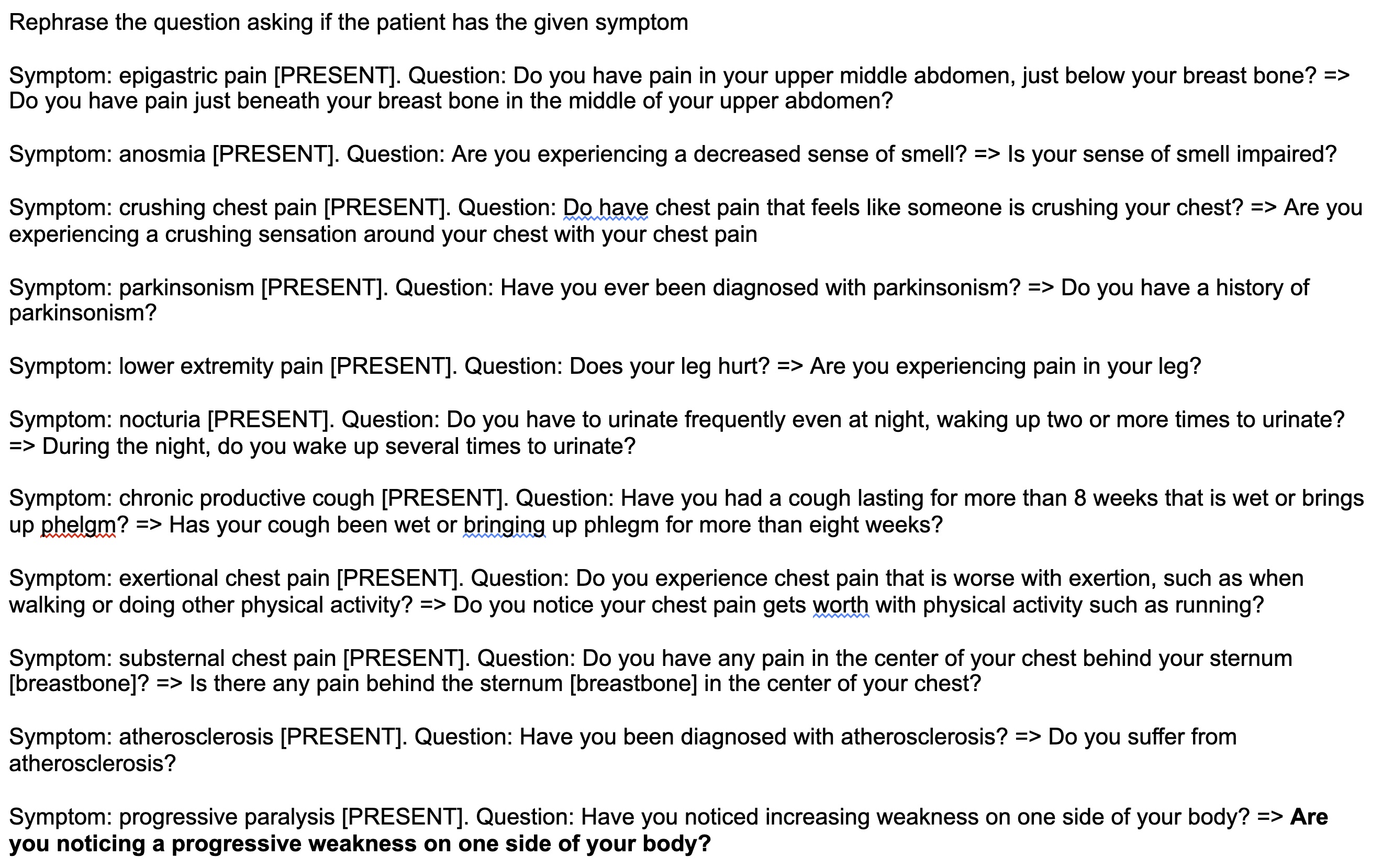}}%
    \qquad
    \subfigure[Second invocation]{
      \includegraphics[width=\linewidth]{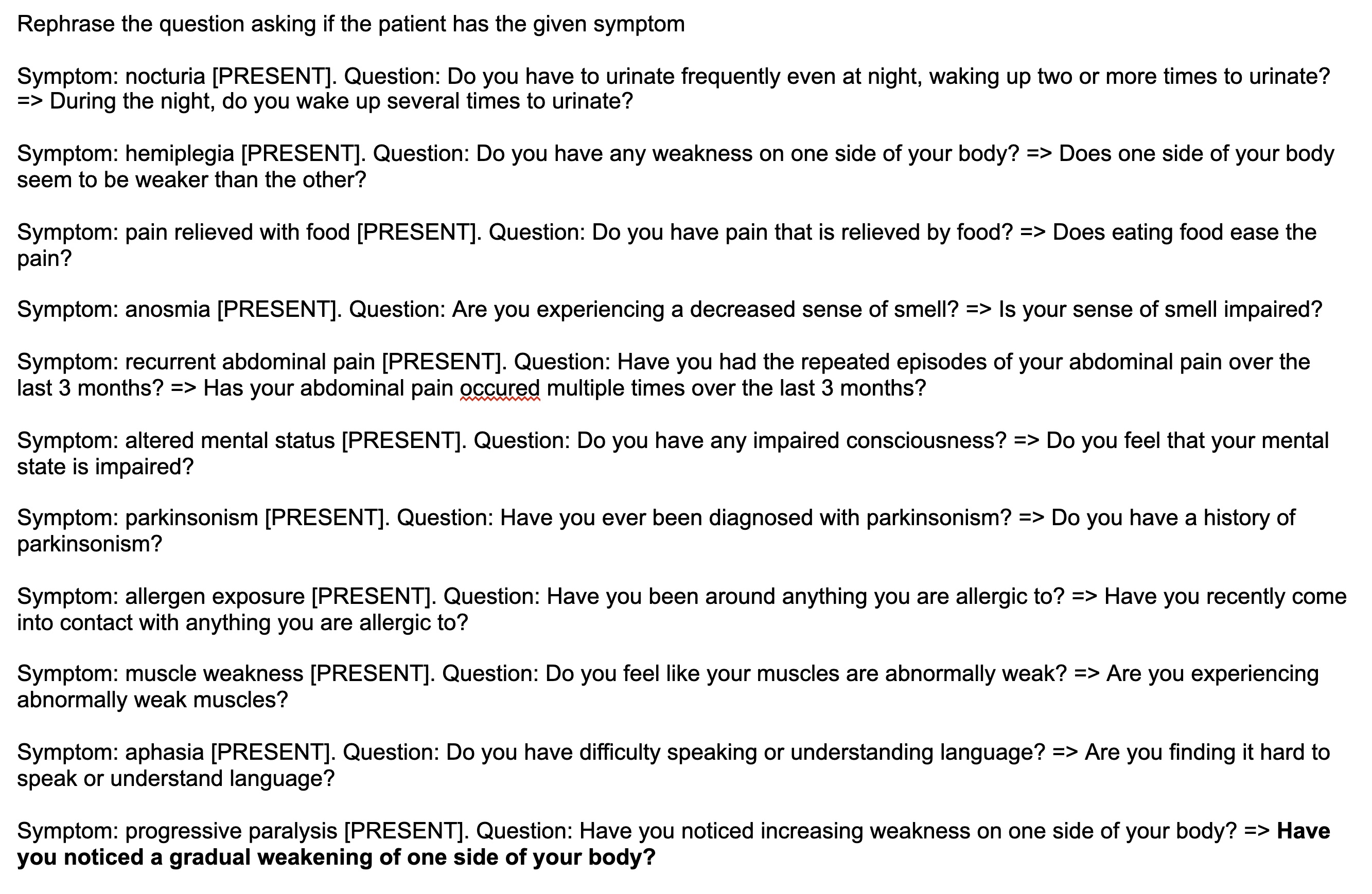}}
   }
\end{figure*}
\end{document}